\documentclass[letterpaper, 10 pt, journal, twoside]{IEEEtran}
\IEEEpubid{\parbox{\textwidth}{\vspace{48pt} \copyright \ 2025 IEEE. Personal use of this material is permitted. Permission from IEEE must be obtained for all other uses, in any current or future media, including reprinting/republishing this material for advertising or promotional purposes, creating new collective works, for resale or redistribution to servers or lists, or reuse of any copyrighted component of this work in other works.}}
\usepackage{graphics} % for pdf, bitmapped graphics files
\usepackage{times} % assumes new font selection scheme installed
\usepackage{amsmath} % assumes amsmath package installed
\usepackage{amssymb}  % assumes amsmath package installed
\usepackage{lipsum}
\usepackage{graphicx}
\usepackage{makecell}
\usepackage{multirow}
\usepackage{tabularx}
\usepackage{float}
\usepackage{cite}
\usepackage{rotating}
\usepackage{threeparttable}
\usepackage{url}
\usepackage[table,dvipsnames]{xcolor}

\newcommand\rotmulti[1]{\rotatebox{90}{\parbox{1.6cm}{#1}}}
\newcommand\rott[1]{\rotatebox{90}{\parbox{1.8cm}{#1}}}
\newcommand\rotts[1]{\rotatebox{90}{\parbox{1.7cm}{#1}}}
\newcommand\rots[1]{\rotatebox{90}{\parbox{1.1cm}{#1}}}
\newcommand{\tb}[1]{\textbf{#1}}
\newcommand{\tu}[1]{\underline{#1}}

\begin{document}

\title{
DINO-VO: A Feature-based Visual Odometry Leveraging a Visual Foundation Model
}

\author{Maulana Bisyir Azhari$^{1}$ and David Hyunchul Shim$^{1}$ % <-this % stops a space

\thanks{Manuscript received: March, 25, 2025; Revised June, 12, 2025; Accepted July, 13, 2025.}%Use only for final RAL version
\thanks{This paper was recommended for publication by Editor Sven Behnke upon evaluation of the Associate Editor and Reviewers' comments.
This work was financially supported by the Institute of Civil Military Technology Cooperation funded by the Defense Acquisition Program Administration and Ministry of Trade, Industry, and Energy of Korean goverment under grant No. UM222006RD2. (\textit{Corresponding author: David Hyunchul Shim.})} %Use only for final RAL version
% \thanks{$^{*}$Corresponding author.}
\thanks{$^{1}$The authors are with the School of Electrical Engineering, Korea Advanced
Institute of Science and Technology, Daejeon 305701, South Korea (e-mail: {\tt\footnotesize mbazhari@kaist.ac.kr; hcshim@kaist.ac.kr}).}%

\thanks{Digital Object Identifier (DOI): see top of this page.}

}

% \markboth{IEEE Robotics and Automation Letters. Preprint Version. Accepted July, 2025}
% {Azhari \MakeLowercase{\textit{et al.}}: DINO-VO: A Feature-based Visual Odometry Leveraging a Visual Foundation Model}

\markboth{IEEE Robotics and Automation Letters. Preprint Version. Accepted July, 2025}
{Azhari \MakeLowercase{\textit{et al.}}: DINO-VO: A Feature-based Visual Odometry Leveraging a Visual Foundation Model}



\maketitle
% \thispagestyle{empty}
% \pagestyle{empty}

% \vspace{-32pt}
%%%%%%%%%%%%%%%%%%%%%%%%%%%%%%%%%%%%%%%%%%%%%%%%%%%%%%%%%%%%%%%%%%%%%%%%%%%%%%%%
\begin{abstract}
Learning-based monocular visual odometry (VO) poses robustness, generalization, and efficiency challenges in robotics.
Recent advances in visual foundation models, such as DINOv2, have improved robustness and generalization in various vision tasks, yet their integration in VO remains limited due to coarse feature granularity.
In this paper, we present DINO-VO, a feature-based VO system leveraging DINOv2 visual foundation model for its sparse feature matching.
To address the integration challenge, we propose a salient keypoints detector tailored to DINOv2’s coarse features.
Furthermore, we complement DINOv2's robust-semantic features with fine-grained geometric features, resulting in more localizable representations.
Finally, a transformer-based matcher and differentiable pose estimation layer enable precise camera motion estimation by learning good matches. 
Against prior detector-descriptor networks like SuperPoint, DINO-VO demonstrates greater robustness in challenging environments.
Furthermore, we show superior accuracy and generalization of the proposed feature descriptors against standalone DINOv2 coarse features.
DINO-VO outperforms prior frame-to-frame VO methods on the TartanAir and KITTI datasets and is competitive on EuRoC dataset, while running efficiently at 72 FPS with less than 1GB of memory usage on a single GPU.
Moreover, it performs competitively against Visual SLAM systems on outdoor driving scenarios, showcasing its generalization capabilities.
\end{abstract}


% \IEEEpeerreviewmaketitle

%%%%%%%%%%%%%%%%%%%%%%%%%%%%%%%%%%%%%%%%%%%%%%%%%%%%%%%%%%%%%%%%%%%%%%%%%%%%%%%%
\vspace{-8pt}
\section{Introduction}
\label{sec:intro}
\IEEEPARstart{V}{isual odometry (VO)} is a technique used to estimate the motion of a camera or robot by analyzing changes in visual data captured from consecutive images. 
VO is a critical component of many modern technologies, including autonomous vehicles, drones, augmented reality, and robotics. 
However, acquiring robust and accurate VO is challenging in environments with low visual texture, poor lighting, rapid motion, and dynamic objects.
Traditional VO methods extract a set of keypoints using handcrafted features~\cite{campos2021orb3} or directly from pixel intensities~\cite{engel2017dso}, associate them in successive frames, recover the camera motion, and optimize the poses and structures. 
However, they lack robustness in low-texture environments, under variable lighting, and in dynamic object scenarios.
Recent works have aimed to improve robustness by utilizing learning-based features~\cite{tang2019gcnv2, ilter2024semantically} or via end-to-end learning frameworks~\cite{wang2017deepvo, teed2018deepv2d, wang2021tartanvo, teed2021droid, teed2024dpvo, lipson2024dpvslam} with auxiliary objectives such as depth and/or optical flow.
Nonetheless, these methods struggle with generalization~\cite{tang2019gcnv2, ilter2024semantically, wang2017deepvo, teed2018deepv2d} and are memory-intensive~\cite{teed2021droid, teed2024dpvo, lipson2024dpvslam}.

\begin{figure}[t!]
\begin{center}
\includegraphics[width=1\columnwidth]{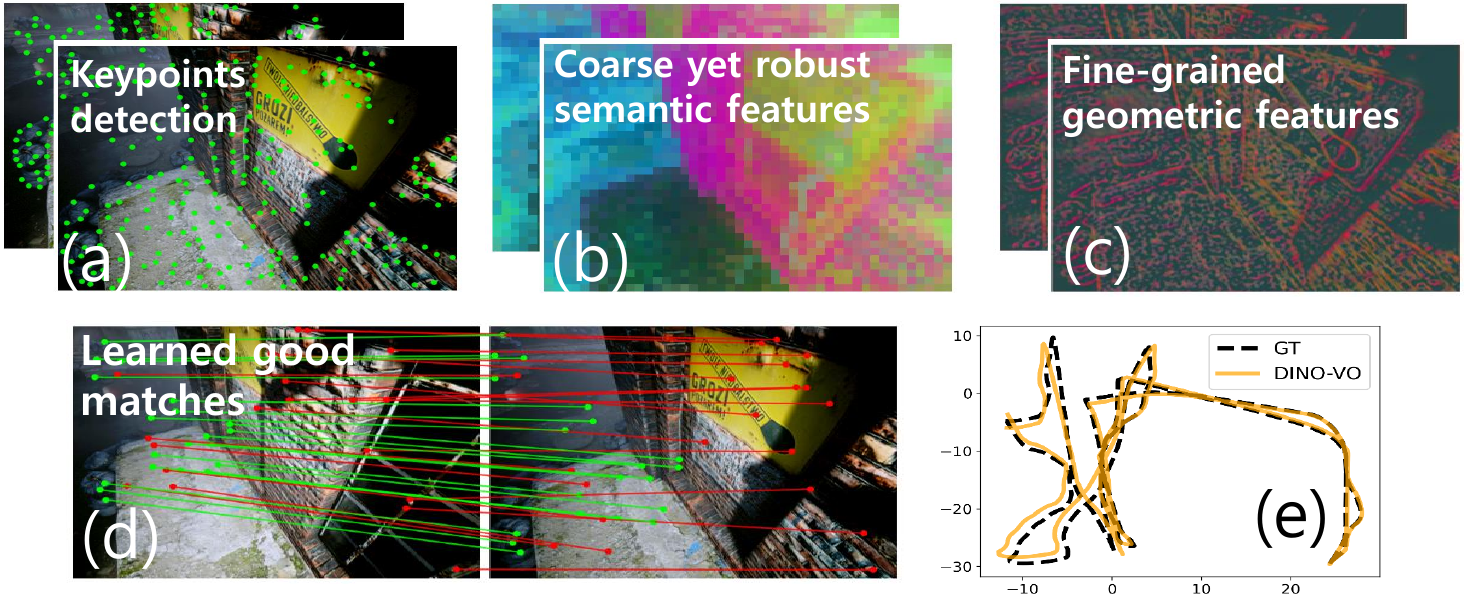}
\end{center}
\vspace{-14pt}
\caption{High-level overview of DINO-VO. (a) The proposed keypoints detection query its corresponding (b) coarse yet robust semantic features from visual foundation model and (c) fine-grained localizable geometric features. (d) The keypoints are matched along with their predicted confidences (showing the \textcolor{green}{20 highest} and \textcolor{red}{20 lowest}-confidence matches). (e) Trajectory estimation on TartanAir dataset sequence MH000, where DINO-VO tracks the ego-motion in a frame-to-frame manner with low-drift.}
\vspace{-12pt}
\label{fig:high_level}
\end{figure}

Recently, visual foundation models~\cite{caron2021dino, oquab2024dinov2} have demonstrated promising results in improving both robustness and generalization across various tasks. 
Specifically, DINOv2~\cite{oquab2024dinov2}, a strong pre-trained image encoder based on Vision Transformer (ViT)~\cite{dosovitskiy2021vit}, has been demonstrated to perform well in other vision tasks~\cite{yang2024depth, edstedt2024roma, jiang2024omniglue, tumanyan2024dinotracker, keetha2023anyloc}.
DINOv2 has the potential of improving the robustness and generalization ability in VO.
However, the direct application of DINOv2 on VO is not straightforward, as the ViT encoder results in coarse features due to patchification~\cite{edstedt2024roma, tumanyan2024dinotracker}.
\begin{figure*}[t!]
\begin{center}
\includegraphics[width=1.0\textwidth]{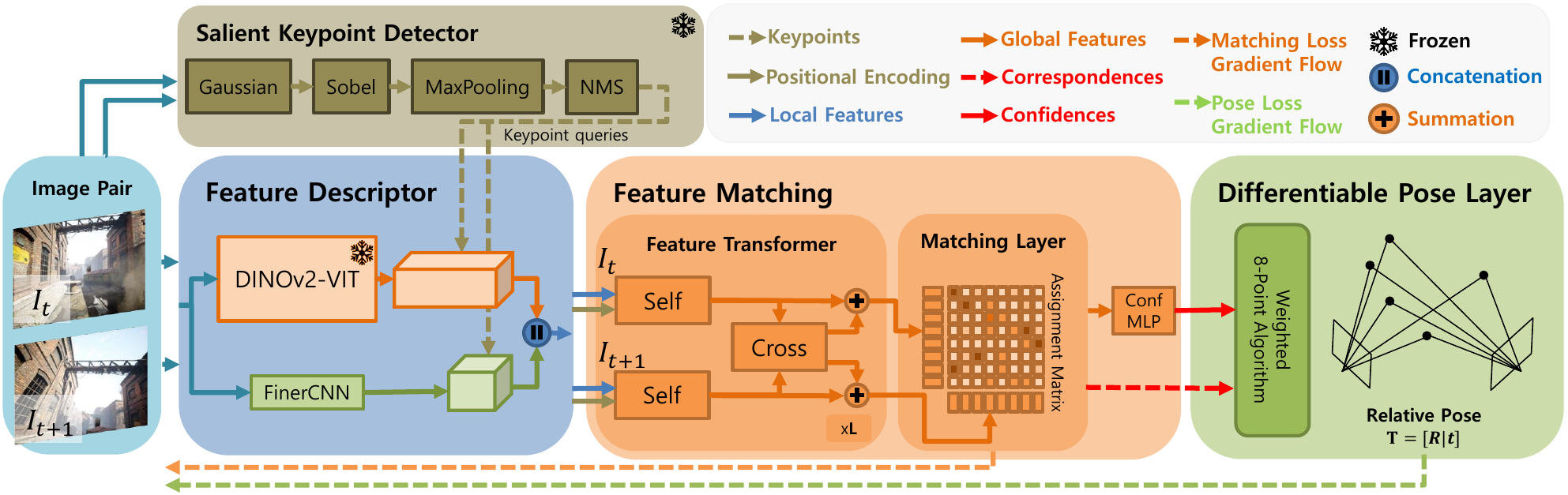}
\vspace{-20pt}
\caption{Overall architecture of the proposed method. The visual odometry (VO) system receives a pair of images $\{I_t, I_{t+1}\}$ and predict its relative transformation $\mathbf{T}=[\mathbf{R}|\mathbf{t}]$. The proposed method consists of 4 modules. First, the keypoints detector detects salient keypoints of each image that align with the coarseness of the DINOv2-ViT feature descriptor. Second, the keypoints query its feature descriptor generated by DINOv2-ViT Encoder --which provides robust and generalizable features-- accompanied by a lightweight CNN Encoder to provide low-level geometrical and localizable features. Third, a transformer-based feature matching network is employed to get keypoint correspondences along with its confidence. Lasty, a differentiable pose layer predicts the image's transformation given its keypoint correspondences and confidences.}
\vspace{-16pt}
\label{fig/proposed_method}
\end{center}
\end{figure*}

In this paper, we introduce \textbf{DINO-VO}, a novel approach that leverages DINOv2 visual foundation model~\cite{oquab2024dinov2} to improve robustness and generalization for a sparse feature-based VO.
The high-level overview of DINO-VO is depicted in Fig.~\ref{fig:high_level}.
To address the issue of coarse features, we propose a lightweight keypoints detector specifically designed to align with DINOv2's coarseness.
We further employ a lightweight CNN encoder to enhance keypoint localization by providing fine-grained, low-level geometric features.
Furthermore, we utilize transformer-based feature matching \cite{lindenberger2023lightglue} to find keypoint correspondences. 
The utilization of learning-based matching layers, particularly transformer-based methods, has proven highly beneficial and demonstrated strong performance~\cite{lindenberger2023lightglue, wang2024efficientloftr, edstedt2024roma}.
When coupled with a learned differentiable pose estimation layer \cite{roessle2023end2endmatching}, this approach allows us to focus directly on learning good matches for estimating camera motion.
% \vspace{8pt}

To summarize, our contributions are as follows:
\begin{itemize}
    \item DINO-VO is a novel VO framework that directly leverages DINOv2~\cite{oquab2024dinov2} visual foundation model features resulting in a robust and general system. 
    \item We propose a salient keypoints detector and a fine-grained geometric descriptor to complement DINOv2 coarse features, which are lightweight yet significantly improve the VO performance.
    \item We validate DINO-VO on TartanAir, EuRoC, and KITTI datasets, achieving state-of-the-art performances compared to prior frame-to-frame VO methods. We further show the efficacy of our design choices on various sequences.
\end{itemize}

\vspace{-4pt}
\section{Related Work}
\label{sec:related_works}
\subsection{Learned Monocular Visual Odometry}
The first notable learning-based VO is DeepVO\cite{wang2017deepvo}, which presents an end-to-end VO using deep RCNNs. 
D3VO\cite{yang2020d3vo} integrates depth, pose, and uncertainty estimations in its sparse direct visual odometry framework.
TartanVO~\cite{wang2021tartanvo} and iSLAM~\cite{fu2024islam} estimate camera motion using optical flow prediction and an intrinsics camera layer, enabling out-of-domain evaluation and real-world use. 
DiffposeNet\cite{parameshwara2022diffposenet} predicts normal flow for pose estimation via a differentiable cheirality layer. 
XVO~\cite{lai2023xvo} utilizes multimodal supervision to achieve generalization across diverse driving scenarios.
Other methods such as DROID-SLAM~\cite{teed2021droid}, DPVO~\cite{teed2024dpvo} and DPV-SLAM~\cite{lipson2024dpvslam} integrate RAFT \cite{teed2020raft} in an end-to-end system combining optical flow and uncertainty estimation using iterative GRU updates, followed by geometric optimization via bundle adjustment. 
However, these systems~\cite{teed2021droid, teed2024dpvo, lipson2024dpvslam} mostly rely on dense optical flow prediction which requires intensive memory and/or computation. Despite the effort of using "sparse patch" in their optical flow matching processes, \cite{teed2024dpvo} and \cite{lipson2024dpvslam} will still require storing the dense feature map for the optical-flow matching process.

\begin{figure}[t!]
\begin{center}
\includegraphics[width=1.0\columnwidth]{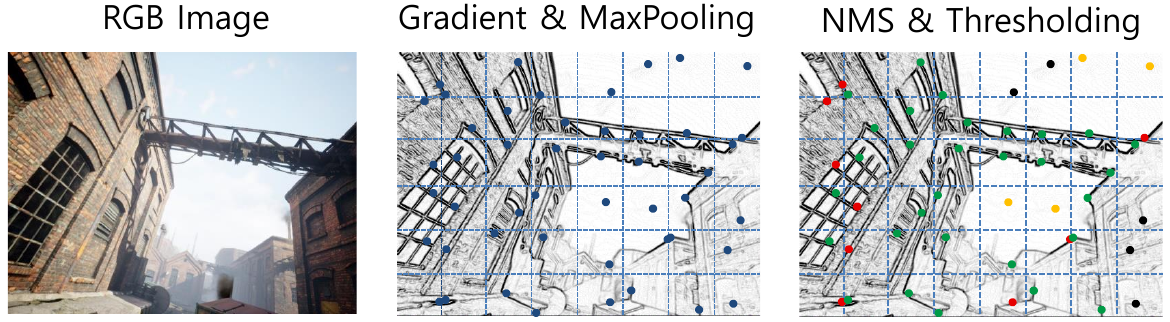}
\end{center}
\vspace{-14pt}
\caption{Visualization of salient keypoints detector. \textbf{left}: input image. \textbf{middle}: gradient map and salient point selection on each grid, shown in blue. \textbf{right}: the red, orange, and black points correspond to removed points after NMS, thresholding, and top-k selections, respectively. The green points are the chosen keypoints and will be used to query feature descriptors.}
\vspace{-12pt}
\label{fig/keypoint_det}
\end{figure}

\subsection{Foundation Models for Computer Vision Tasks} 
Foundation models for computer vision have advanced rapidly in recent years~\cite{caron2021dino,oquab2024dinov2}. 
DINOv2~\cite{oquab2024dinov2}, in particular, introduced a scalable self-supervised approach that produces high-quality visual features without labeled data, achieving state-of-the-art results across various perception tasks~\cite{yang2024depth, edstedt2024roma, jiang2024omniglue,tumanyan2024dinotracker, keetha2023anyloc}.
RoMa~\cite{edstedt2024roma} utilized DINOv2 for coarse matching in a two-stage dense image matching task, refined by a specialized ConvNet, achieving state-of-the-art performance across multiple datasets. 
Similarly, Dino-tracker~\cite{tumanyan2024dinotracker} used DINOv2 for a dense pixel-tracking task.
MambaVO\cite{wang2025mambavo} uses DINOv2 features in its demi-dense matching which are then refined using the Geometric Mamba Module.
In sparse matching task, DINOv2 guided the matching layer in OmniGlue \cite{jiang2024omniglue}, significantly improving zero-shot generalization. 
However, DINOv2's coarse feature map complicates its integration into tasks requiring pixel-level accuracy, such as visual odometry. 
Unlike previous approaches \cite{edstedt2024roma, jiang2024omniglue, wang2025mambavo, tumanyan2024dinotracker}, we aim to use DINOv2 directly in one-stage sparse image matching to meet the real-time requirement of visual odometry.

\vspace{-4pt}
\section{Method}
\label{sec:proposed}
The objective of our work is to predict the relative transformation $\mathbf{T}= [\mathbf{R}|\mathbf{t}]$, composed of rotation matrix $\mathbf{R} \in SO(3)$ and up-to-scale translation vector $\mathbf{t} \in \mathbb{R}^3$, between two consecutive images $\{I_t,I_{t+1}\} \in \mathbb{R}^{H \times W \times 3}$ for a VO frontend system. 
As shown in Fig.~\ref{fig/proposed_method}, our proposed VO system integrates four key components: a salient keypoints detector, a feature descriptor networks that utilize DINOv2 along with a lightweight CNN encoder, a transformer-based feature matching layer, and a differentiable pose layer.

\subsection{Salient Keypoints Detector}
\label{sec:keypoint_det}
As noted in\cite{jiang2024omniglue}, combining SuperPoint\cite{detone2018superpoint} with DINOv2’s coarse features poses challenges in sparse matching. 
RoMa~\cite{edstedt2024roma} showed DINOv2 lacks precise localization, prompting two stage dense matching, but its high computational cost limits real-time VO. 
Our proposed keypoints detector queries DINOv2 features from sparse salient points, enabling precise, one-stage matching with low computational demands.
The visualization of the proposed keypoints detector is depicted in Fig.~\ref{fig/keypoint_det}.

The keypoints detector computes the gradient map using a $\operatorname{GaussianFilter}$ and a $\operatorname{SobelFilter}$.
The gradient map is then divided into $H/r_{\text{P}} \times W/r_{\text{P}}$ grids, aligning keypoints with DINOv2 coarse features, where $r_{\text{P}}=14$ corresponds to the patch size of the DINOv2-ViT.
We select the point with the highest gradient magnitude in each grid using a $\operatorname{MaxPooling}$ operator with a kernel size of $r_{\text{P}}$ and a stride of $r_{\text{P}}$, resulting in the most salient point in each grid (shown in \textcolor{blue}{blue)}.
Oftentimes, the salient points are located at the grid borders, leading to keypoints that are very close to one another, making them redundant (shown in \textcolor{red}{red}).
To address this, we apply $\operatorname{NonMaxSuppression}$ ($\operatorname{NMS}$)~\cite{detone2018superpoint} with a radius of $r_{\operatorname{NMS}}$.
In addition to removing redundant points, a large $r_{\operatorname{NMS}}$ ensures that keypoints are well-distributed across the image, which improves the robustness and accuracy of pose estimation.
Finally, we filter out keypoints with very small gradient magnitudes (shown in \textcolor{orange}{orange}) and select the top-\textit{k} keypoints based on their gradient magnitudes (shown in \textcolor{green}{green}).
The output, keypoint locations $ \mathcal{K} = \{ (x_i, y_i) \}_{i=1}^{K} $ where $ x_i \in [1,H] $ and $ y_i \in [1, W] $, is then used to query features described in the next subsection.
\begin{figure}[t!]
\begin{center}
\includegraphics[width=1.0\columnwidth]{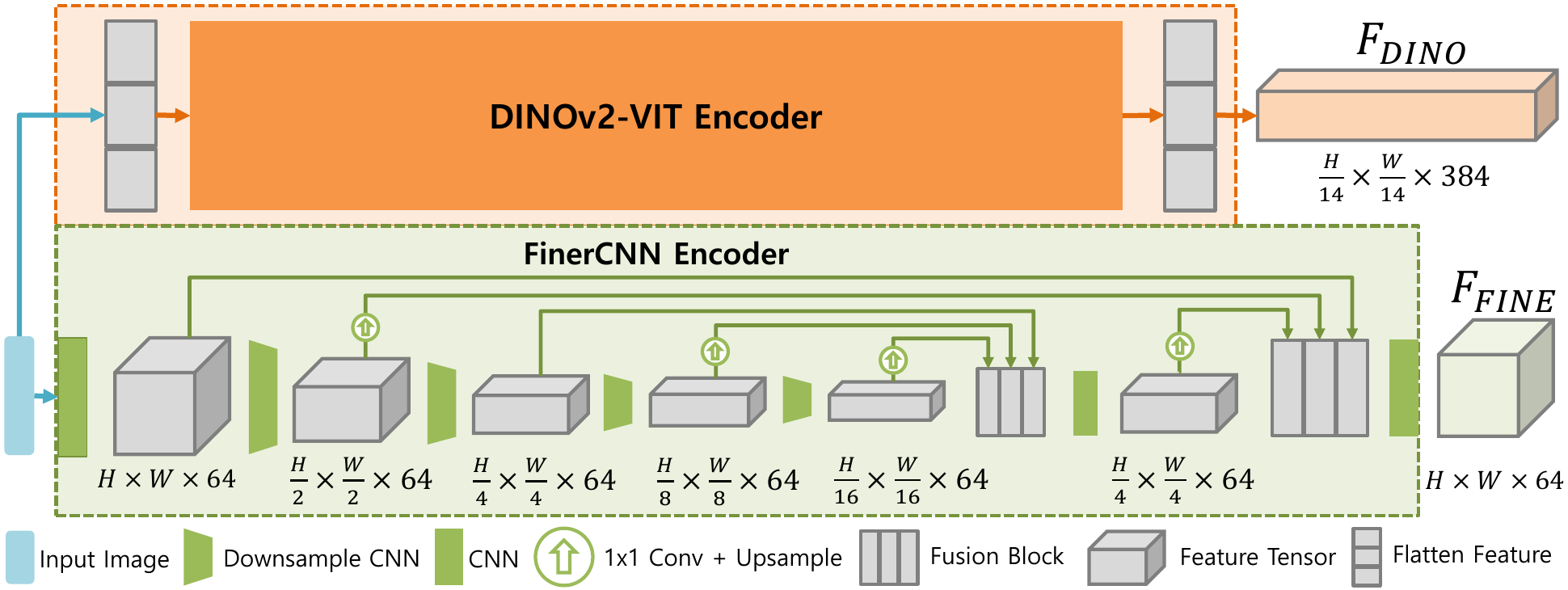}
\end{center}
\vspace{-0.2cm}
\caption{Feature descriptor networks. It consists of two encoders: a DINOv2-ViT and a specialized lightweight CNN encoder, FinerCNN.}
\vspace{-0.5cm}
\label{fig/feature_desc}
\end{figure}
\subsection{Feature Descriptor} 
\label{sec:feat_desc}

As shown in Fig.~\ref{fig/feature_desc}, the proposed feature descriptor consists of two networks: DINOv2 and FinerCNN.
The DINOv2 encoder extracts generalizable and robust semantic features for our matching process and produces a feature map $\mathbf{F}_{\text{DINO}} \in \mathbb{R}^{\frac{H}{14} \times \frac{W}{14} \times 384}$.
To complement DINOv2 coarse features, we employ FinerCNN to provide fine-grained low-level geometric features.
FinerCNN is built from CNN blocks called basic layers, inspired by XFeat~\cite{potje2024xfeat}, which has shown great performance in image matching while having lightweight design. 
We adopt the feature pyramid approach to expand the receptive field until the resolution is reduced to $H/16 \times W/16$. 
The intermediate features are then fused at $H/4 \times W/4$ and $H \times W$ resolution levels as shown in Fig.~\ref{fig/feature_desc}. 
FinerCNN generates a feature map $\mathbf{F}_{\text{FINE}} \in \mathbb{R}^{H \times W \times 64}$ that has the same spatial dimension as the input image providing the optimal localizable features at pixel-level resolution.

For each keypoint location $ (x_i, y_i) \in \mathcal{K} $, we directly retrieve its corresponding feature from the feature map by indexing it as $\mathbf{f}_{\text{FINE}}^i := \mathbf{F}_{\text{FINE}}(x_i, y_i, :) \in \mathbb{R}^{64} $.
To query its corresponding feature in $\mathbf{F}_{\text{DINO}} $, we apply division and floor operations to the keypoint location $ \mathbf{f}_{\text{DINO}}^i := \mathbf{F}_{\text{DINO}}(\lfloor \frac{x_i}{14} \rfloor, \lfloor \frac{y_i}{14} \rfloor, :) \in \mathbb{R}^{384} $. 
Finally, the two features are concatenated, and a $\operatorname{Linear}$ projector is used to reduce the dimensionality.
\begin{equation}
\label{eq:concat_features}
    \mathbf{f}_i= \operatorname{Linear} ([\mathbf{f}_{\text{DINO}}^i | \mathbf{f}_{\text{FINE}}^i]) \in \mathbb{R} ^{192} \enspace .
\end{equation}
where $[\cdot|\cdot]$ concatenates two vectors in the channel dimension.

\subsection{Feature Matching}
Given two consecutive images $ I_t $ and $ I_{t+1} $, we extract $ K $ features $ \mathbf{f}_i^{I_{t}} \in \mathbb{R}^{192} $ indexed by $\mathcal{K}_t=\{(x_i,y_i)\}_{i=1}^K$ for image $ I_t $ and $ \mathbf{f}_j^{I_{t+1}} \in \mathbb{R}^{192} $ indexed by by $\mathcal{K}_{t+1}=\{(x_j,y_j)\}_{j=1}^K$ for image $ I_{t+1} $, using feature descriptor explained in previous subsection.
Features from the two images are matched by predicting a partial assignment $ \mathcal{M} = \{(i,j)\} \subset \mathcal{K}_t \times \mathcal{K}_{t+1} $.
We use learnable feature matching based on Transformer architecture, as it has been shown to be more robust compared to traditional feature matching methods~\cite{sarlin2020superglue, lindenberger2023lightglue, jiang2024omniglue}.
Specifically, following~\cite{lindenberger2023lightglue}, we use $ L $ identical transformer layers consisting of self-attention and cross-attention units to process the local features into features with global context.
In each attention unit, the feature descriptor $ \mathbf{f}_i^T$ is updated using the following formulas:
\begin{align}
    \mathbf{f}_i^T \leftarrow \mathbf{f}_i^T + \text{MLP} ([\mathbf{f}_i^T|\mathbf{m}_i^{T \leftarrow S}]), \enspace \text{with} \\
    \mathbf{m}^{T\leftarrow S}_i = \sum_{j\in\mathcal{S}}{\operatorname{Softmax}}\left(a_{ij}^{TS}\right)_j
    \mathbf{v}_j, \enspace \text{and} \\
    a_{ij}^{TS} = 
    \begin{cases}
        \mathbf{q}_i^\top\,\mathbf{R_P}\!\left(\mathbf{x}_j - \mathbf{x}_i\right)\,\mathbf{k}_j, &\text{for self-attention}\\ 
        \mathbf{k}_i^T{}^\top\mathbf{k}_j^S \overset{!}{=} a_{ji}^{ST}, &\text{for cross-attention} \\
    \end{cases}
    \label{eq:attention_score}
\end{align}
where $\mathbf{m}_i^{T \leftarrow S}$ is the aggregated message from a source image $S \in \{I_t, I_{t+1}\}$ to a target image $T \in \{I_t, I_{t+1}\}$ using attention mechanism.
$a_{ij}^{TS}$ is the attention score between feature $ i $ of image $ T $ and feature $ j $ of image $ S $ which is computed differently in the self- and cross-attention unit using Eq. \ref{eq:attention_score}, with $S=T$ in the self-attention and $S=\{I_t,I_{t+1}\}\backslash T$ in the cross-attention. 
$\mathbf{k}_i$, $\mathbf{q}_i$, and $\mathbf{v}_i$ are the key, query and value vector of linearly projected $\mathbf{f}_i$, respectively.
$\mathbf{R_P}(\cdot) \in \mathbb{R}^{d \times d}$ is a rotary positional encoding \cite{su2024roformer} and $\mathbf{p}_i:=(x_i,y_i) \in [0,1]^2$ is the normalized keypoint location of feature $i$.

The point correspondences can be obtained by combining the feature similarity $\mathbf{S}_{ij}$ and feature-wise matchability scores $\sigma_i$, $\sigma_j$ into a soft partial assignment matrix $\mathbf{P} \in [0,1]^{K \times K}$ as
\begin{align}
    \mathbf{P}_{ij} =
    \sigma_i\;
    \sigma_j\;
    \underset{k\in\mathcal{K}_t}{\operatorname{Softmax}}(\mathbf{S}_{kj})_i\;
    \underset{k\in\mathcal{K}_{t+1}}{\operatorname{Softmax}}(\mathbf{S}_{ik})_j, \enspace \text{where} \\
    \mathbf{S}_{ij} = \operatorname{Linear}\left(\mathbf{f}_i^{I_{t}}\right){}^\top\operatorname{Linear}\left(\mathbf{f}_j^{I_{t+1}}\right), \enspace\text{and} \\
    \sigma_i = \operatorname{Sigmoid}\left(\operatorname{Linear}(\mathbf{f}_i^{I_{t}})\right) \enspace \in[0,1], \\
    \sigma_j = \operatorname{Sigmoid}\left(\operatorname{Linear}(\mathbf{f}_j^{I_{t+1}})\right)\in[0,1],
\end{align}
A pair of points $(i,j)$ yields a correspondence when both points are predicted as matchable and when their similarity is higher than any other point in both images. 
Furthermore, for each pair of matching points $(i,j)$ a confidence weight $w_{ij}$ is predicted from the final layer descriptor $\mathbf{f}_i^{I_{t}}$ and $\mathbf{f}_j^{I_{t+1}}$ as:
\begin{equation}
w_{ij} = \operatorname{ConfMLP}\left(\left[\mathbf{f}_i^{I_{t}} | \mathbf{f}_j^{I_{t+1}}\right]\right) \enspace.
\label{eq:weight_mlp}
\end{equation}

\subsection{Differentiable Pose Estimation}
\label{sec:eight_point}
We utilize a confidence-weighted eight-point algorithm as the differentiable pose layer to estimate the relative pose between two images given its corresponding points. 
The weighted eight-point algorithm has shown to be useful and efficient for providing supervision to the matching layer to learn good matches for pose estimation\cite{roessle2023end2endmatching}.
For each pair of corresponding points $\mathbf{x}_i^{\mathcal{K}_{t}} = [x_i, y_i, 1]^T$ and $\mathbf{x}_j^{\mathcal{K}_{t+1}} = [x_j, y_j, 1]^T$ with $x,y \in [0,1]$, the essential matrix $\mathbf{E} \in \mathbb{R}^{3 \times 3}$ satisfies the following equation:
\begin{equation}
\mathbf{x}_j^T \mathbf{E} \mathbf{x}_i = 0 \enspace,
\end{equation}

By stacking the known coordinates of correspondences into a matrix $\mathbf{\Phi}$, where each row $[x_i x_j, x_i y_j, x, y_ix_j, y_iy_j, y, x_j, y_j, 1]$ describes one correspondence and flattening $\mathbf{E}$ into $\operatorname{flat}(\mathbf{E})=[e_{11}, e_{12}, e_{13}, e_{21}, e_{22}, e_{23}, e_{31}, e_{32}, e_{33}]^T$, we can rewrite the confidence-weighted version of the eight-point algorithm into a matrix equation
\begin{equation}
\operatorname{diag}(\mathbf{w})\mathbf{\Phi}\operatorname{flat}(\mathbf{E}) = \mathbf{0} \enspace,
\label{eq:e_matrix}
\end{equation}
where $\mathbf{w}$ is the weight vector obtained by stacking the corresponding $w_{ij}$ of every match from (\ref{eq:weight_mlp}).

We solve the Eq. (\ref{eq:e_matrix}) by using Singular value decomposition (SVD) of $\operatorname{diag(\mathbf{w})\mathbf{\Phi}}$, which minimizes $\|\operatorname{diag}(\mathbf{w})\operatorname{flat}(\mathbf{E})\|_2$ and enforcing $\mathbf{E}$ to have rank 2. 
$\mathbf{E}$ can be decomposed into multiple relative transformations $\mathbf{T}$, the best $\mathbf{T}$ is then chosen by using cheirality condition \cite{hartley2003multiple}.

\subsection{Supervision}
Following \cite{lindenberger2023lightglue}, we supervise the assignment matrix $\mathbf{P}$ using ground truth labels derived from two-view transformations and pixel-wise depth. We minimize the log-likelihood of the assignment predicted at each layer $\ell$. The matching objective $\mathcal{L}_{m}$ is defined as:
\begin{equation}
    \begin{split}
    \mathcal{L}_{m} =
     -\frac{1}{L}\sum_{\ell}\Bigg(
     &\frac{1}{|\mathcal{M}|}\sum_{(i, j) \in \mathcal{M}} \log {}^\ell\*P_{ij}\\
     &+ \frac{1}{2|\bar{\mathcal{K}}_t|}\sum_{i \in \bar{\mathcal{K}}_t} \log \left(1 - {}^\ell\sigma_i^{\mathcal{K}_t}\right)\\
     &+ \frac{1}{2|\bar{\mathcal{K}}_{t+1}|}\sum_{j \in \bar{\mathcal{K}}_{t+1}} \log \left(1 - {}^\ell\sigma_j^{\mathcal{K}_{t+1}}\right)\Bigg)\enspace.
\end{split}
\end{equation}
where $\bar{\mathcal{K}}_t\subseteq \mathcal{K}_t$ and $\bar{\mathcal{K}}_{t+1}\subseteq \mathcal{K}_{t+1}$ are unmatchable keypoints.

We also minimize the difference between the predicted transformation $\hat{\mathbf{T}}=[\hat{\mathbf{R}}|\hat{\mathbf{t}}] \in Sim(3)$ and the ground-truth $\mathbf{T}=[\mathbf{R}|\mathbf{t}] \in SE(3)$ as the up-to-scale pose objective $\mathcal{L}_{p}$ following \cite{wang2021tartanvo} to handle scale ambiguity in monocular setting, which is defined as:
\begin{equation}
\begin{split}
    \mathcal{L}_{\mathrm{p}}= \enspace &\lambda_{\mathrm{t}} \left \| \frac{\hat{\mathbf{t}}}{\operatorname{max}(||\hat{\mathbf{t}}||, \epsilon)} - \frac{\mathbf{t}}{\operatorname{max}(||\mathbf{t}||, \epsilon)} \right \| \\
    &+ \lambda_{\mathrm{r}} \|\operatorname{Log}(\hat{\mathbf{R}})-\operatorname{Log}(\mathbf{R})\| \enspace.
\end{split}
\end{equation}
where $\epsilon=1e^{-6}$ to stabilize the numerical division, $\lambda_{\mathrm{t}}$ and $\lambda_{\mathrm{r}}$ are the weighting factor for the translation and rotation objective, respectively. $\operatorname{Log}(\cdot)$ is a mapping function from $SO(3)$ to $so(3)$.
The pose objective guides the matching layer to down-weight matches that are detrimental to pose accuracy.
The overall objective $\mathcal{L}_{\mathrm{t}}$ is defined as a weighted summation of both objectives:
\begin{equation}
\label{eq:loss_total}
\mathcal{L}_{\mathrm{t}}=(1-\lambda_{p})\mathcal{L}_{m} + \lambda_{p}\mathcal{L}_{p}.
\end{equation}
where $=\lambda_{\mathrm{p}}$ is the weighting factor of the pose objective.

\subsection{DINO-VO}
DINO-VO predicts the relative pose to the latest keyframe, which is selected when the mean pixel displacement of predicted correspondences exceeds 24px.
This ensures sufficient parallax, thereby mitigating instabilities from pure rotation or minimal translation of the weighted 8-point algorithm \cite{zhan2020dfvo}.
Furthermore, alternate frames are processed for EuRoC and KITTI datasets, this step is unnecessary for the TartanAir MH dataset due to its inherently significant camera motion. 
All frame's poses are saved relative to its previous keyframe such that the full camera trajectory is obtained. 

Since DINO-VO only predicts up-to-scale transformation, we follow \cite{wang2021tartanvo} to scale the predicted translation $\mathbf{t}\in \mathbb{R}^3$ with the ground-truth.
DINO-VO uses the image resolution of 476$\times$630, 476$\times$742, and 364$\times$1008 for TartanAir, EuRoC, and KITTI sequences, respectively.

\section{Experiments}
\label{sec:experiments}

\subsection{Implementation Details}

\subsubsection{Detector and Networks Details}
We use $\operatorname{GaussianFilter}$ with kernel size 5 and $std=2.0$, $\operatorname{MaxPooling}$ with kernel size of $r_P=14$ following the patch-size of DINOv2-ViT-s encoder\cite{oquab2024dinov2}, $r_{\operatorname{NMS}}=8$, gradient threshold of $0.01$. 
We select the top 512 keypoints for image matching and pose estimation. 
We utilize LightGlue\cite{lindenberger2023lightglue} as the matching layer with 12 layers, 3 attention heads with 64 head dimensions. 

\subsubsection{Training Procedure} 
We use only TartanAir dataset~\cite{wang2020tartanairdataset} for training. 
We train the model with matching objective $\mathcal{L}_{m}$ for the first 4 epochs and total objective $\mathcal{L}_{t}$ for the next 10 epochs. 
We use $\lambda_{\mathrm{r}}=180$, $\lambda_{\mathrm{t}}=400$, and $\lambda_{\mathrm{p}}=0.0$ up to $\lambda_{\mathrm{p}}=0.9$ with $1.5\times 10^{-4}$ increment per step starting from epoch 5. 
We use a computer equipped with an RTX4090 GPU and an intel i9-12900k CPU with PyTorch. 
The entire training procedure takes about 3 days.

\subsubsection{Metrics} 
Following prior works\cite{teed2024dpvo, wang2021tartanvo}, we report Absolute Trajectory Error (ATE) in meter for TartanAir MH, EuRoC, and KITTI datasets.
Additionally, we report RMSE drift translational $t_{rel}$ (\%) and rotational $r_{rel}$ (deg/100m) on KITTI dataset~\cite{geiger2013visionkitti} sequences 6, 7, 9, and 10.

\begin{table}[t!]
    
    \centering
    \caption{\vspace{-2pt}{ATE[m]$\downarrow$ on the MH Sequences of TartanAir Dataset.}}
    \label{tab:tartan_air_result}
    \vspace{-6pt}
    \begin{tabularx}{\columnwidth}{c|ccccc|ccc}
    \hline
    &\multicolumn{5}{c|}{\tb{Multi-Frame VO}} &\multicolumn{3}{c}{\tb{Frame-to-Frame VO}} \\ 
    \rotmulti{Sequence (MH)} &\rotmulti{SVO~\cite{forster2016svo}} &\rotmulti{DSO\cite{engel2017dso}} &\rotmulti{DeepV2D\cite{teed2018deepv2d}} &\rotmulti{DROID-VO\cite{teed2021droid}} &\rotmulti{DPVO\cite{teed2024dpvo}} &\rotmulti{TartanVO\cite{wang2021tartanvo}} &\rotmulti{DiffPoseNet\cite{parameshwara2022diffposenet}} &\rotts{\tb{Ours\\(DINO-VO)}} \\

    \hline
    %   ORB     SVO DSO     DeepV2D DROID   DPVO    Tartan      DiffPoseNet DINO
    000 &14.42   &9.65   &6.15   &0.32   &0.21   &4.88       &\tu{2.56}  &\tb{0.87}\\
    001 &1.17   &0.35   &2.12   &0.13   &0.04   &\tu{0.26}  &0.31       &\tb{0.19}\\
    002 &4.97   &7.96   &4.54   &0.08   &0.04   &2.00       &\tu{1.57}  &\tb{0.28}\\
    003 &6.88   &3.46   &3.89   &0.09   &0.08   &0.94       &\tu{0.72}  &\tb{0.25}\\
    004 &X      &X      &2.71   &1.52   &0.58   &\tu{1.07}  &\tb{0.82}  &1.19   \\
    005 &19.2   &12.58  &11.55  &0.69   &0.17   &3.19       &\tu{1.83}  &\tb{0.92}\\
    006 &11.27   &8.42   &5.53   &0.39   &0.11   &\tu{1.00}       &1.32       &\tb{0.23}\\
    007 &17.68   &7.50   &3.76   &0.97   &0.15   &2.04       &\tu{1.24}  &\tb{0.73}\\
    \hline
    Avg &X   &X      &5.03   &0.52   &0.17   &1.92       &\tu{1.29}  &\tb{0.58}\\
    \hline
    
    \end{tabularx}
    {\parbox{1.0\columnwidth}{- The best and second best of frame-to-frame VO results are marked as \textbf{bold} and \underline{underlined}.}}
\end{table}

\begin{table}[t!]
    \centering
    \caption{\vspace{-2pt}{ATE[m]$\downarrow$ on the EuRoC MAV Dataset.}}
    \label{tab:euroc_result}
    \vspace{-6pt}
    \setlength{\tabcolsep}{3.4pt}
    \begin{tabularx}{\columnwidth}{c|ccccc|cccc}
    \hline
    &\multicolumn{5}{c|}{\centering \tb{Multi-Frame VO}} &\multicolumn{4}{c}{\centering \tb{Frame-to-Frame VO}} \\ 
    \rotmulti{Sequence} &\rotmulti{SVO~\cite{forster2016svo}} &\rotmulti{DSO\cite{engel2017dso}} &\rotmulti{DeepV2D\cite{teed2018deepv2d}} &\rotmulti{DROID-VO\cite{teed2021droid}} &\rotmulti{DPVO\cite{teed2024dpvo}} &\rotmulti{TartanVO\cite{wang2021tartanvo}} &\rotmulti{iSLAM-VO$^\mp$\cite{fu2024islam}} &\rotmulti{MAC-VO$^\mp$\cite{qiu2025macvo}} &\rotts{\tb{Ours\hspace{0.2cm} (DINO-VO)}} \\
    \hline
    %     SVO     DSO     DeepV2D DROID   DPVO    Tartan      iSLAM-VO    MACVO     DINO
    MH01 &0.100  &0.046  &1.614  &0.163  &0.087  &0.693      &0.320      &\tu{0.240} &\tb{0.150}\\
    MH02 &0.120  &0.046  &1.492  &0.121  &0.055  &0.325      &0.462      &\tu{0.256} &\tb{0.118}\\
    MH03 &0.410  &0.172  &1.635  &0.242  &0.158  &0.550      &0.380     &\tu{0.260} &\tb{0.254}\\
    MH04 &0.430  &3.810  &1.775  &0.399  &0.137  &1.153      &\tu{0.962} &\tb{0.496} &{0.982}\\
    MH05 &0.300  &0.110  &1.013  &0.270  &0.114  &1.021      &\tu{0.500} &0.560 &\tb{0.454}   \\
    \hline
    V101 &0.070  &0.089  &0.717  &0.102  &0.050  &0.447      &{0.366}   &\tb{0.197} &\tu{0.298}\\
    V102 &0.210  &0.107  &0.695  &0.165  &0.140  &0.389     &0.414      &\tb{0.096} &\tu{0.260}\\
    V102 &X      &0.903  &1.483  &0.158  &0.086  &0.622      &\tu{0.313} &\tb{0.198} &{0.559}\\
    \hline
    V201 &0.110  &0.044  &0.839  &0.102  &0.057  &0.433     &0.478      &\tb{0.145} &\tu{0.334}\\
    V202 &0.110  &0.132  &1.052  &0.115  &0.049  &0.749     &0.424      &\tb{0.270} &\tu{0.278}\\
    V202 &1.080  &1.152  &0.592  &0.204  &0.211  &1.152     &1.176      &\tb{0.425} &\tu{0.763}\\
    \hline
    Avg &X      &0.601  &1.173  &0.186  &0.105  &0.680      &{0.527} &\tb{0.286} &\tu{0.404}\\

    \hline
    
    \end{tabularx}
    {\parbox{1.0\columnwidth}{- The best and second best of frame-to-frame VO results are marked as \textbf{bold} and \underline{underlined}.}} \\
    {\parbox{1.0\columnwidth}{- $^\mp$ The method uses stereo camera setup.}}
    \vspace{-0.3cm}
\end{table}

\subsection{Monocular Visual Odometry Results}
In this section, we evaluate the accuracy and robustness of DINO-VO on TartanAir and EuRoC MAV datasets. 
For a fair comparison, we only numerically compare our method with frame-to-frame VO methods such as TartanVO\cite{wang2021tartanvo}, DiffposeNet\cite{parameshwara2022diffposenet}, MAC-VO\cite{qiu2025macvo}, and a stereo VO frontend of iSLAM\cite{fu2024islam}, named iSLAM-VO. 
We also provide the results of the SOTA VO methods which perform multi-frame optimization such as SVO\cite{forster2016svo}, DSO\cite{engel2017dso}, DeepV2D\cite{teed2018deepv2d}, DROID-VO\cite{teed2021droid}, and DPVO\cite{teed2024dpvo}.

In Table \ref{tab:tartan_air_result}, we present the evaluation of DINO-VO on TartanAir MH sequences, which span a variety of simulated environments with numerous challenges, including illumination changes, dynamic objects, low-texture scenes, and significant viewpoint variations—conditions that are challenging for many traditional methods~\cite{wang2020tartanairdataset}. 
Compared to other frame-to-frame VO approaches, DINO-VO demonstrates better accuracy in 7 out of 8 sequences. 
On average, DINO-VO reduces the ATE by 70\% and 55\% compared to TartanVO, and DiffPoseNet, respectively.

In Table \ref{tab:euroc_result}, we evaluate DINO-VO on EuRoC dataset, which is well-known for its challenging sequence for learning-based methods due to lack of training data, challenging lighting conditions, and aggressive motion patterns. 
Compared to other frame-to-frame VO methods, DINO-VO achieves the lowest ATE on 4 out of 11 sequences in the EuRoC dataset, highlighting not only its generalization capabilities but also its accuracy and robustness, as it successfully completes all sequences without abrupt failures. 
DINO-VO reduces the ATE by 40\% and 23\% compared to TartanVO and iSLAM-VO, respectively.
DINO-VO has a higher average ATE compared to MAC-VO, however, MAC-VO requires a stereo camera setup, 6$\times$ GPU memory, and 13$\times$ inference time (see Sec. \ref{sec:runtime_and_memory}).

\begin{table}[t!]
    \centering
    \caption{\vspace{-2pt}{Generalization Assesment Against Other Learned Frame-to-Frame VO methods on KITTI Odometry Sequence 06, 07, 09, and 10 using RMSE Drift$\downarrow$ $t_{rel}$ and $r_{rel}$.}}
    \label{tab:kitti_rpe}

    \vspace{-6pt}
    \setlength{\tabcolsep}{5.45pt}
    \begin{tabularx}{\columnwidth}{cc|c|cccc|c}\hline
    & \multirowcell{2}{Method} &\multirowcell{2}{Met.} &\multicolumn{4}{c|}{Sequence} &\\ 
    & & &06 &07 &09 &10 &Avg \\
    
    \hline

    \multirow{12}{*}{\rotatebox{90}{\parbox{80pt}{\centering Driving Scenarios \hspace{2cm} Frame-to-Frame VO}}}
    &\multirow{2}{*}{DeepVO\cite{wang2017deepvo}}
    &$t_{\text{rel}}$ &5.42 &3.91 &- &8.11 &- \\
    & &$r_{\text{rel}}$ &5.82 &4.60 &- &8.83 &- \\
    \cline{2-8}
    &\multirow{2}{*}{DFVO~\cite{zhan2020dfvo} }
    &$t_{\text{rel}}$ &- &- &2.47 &1.96 &- \\
    & &$r_{\text{rel}}$ &- &- &\tu{\tb{0.30}} &\tu{\tb{0.31}} &- \\
    \cline{2-8}
    &\multirow{2}{*}{HDVO$^\mp$~\cite{liu2022hdvo} }
    &$t_{\text{rel}}$ &- &- &\tu{1.97} &\tu{1.89} &- \\
    & &$r_{\text{rel}}$ &- &- &0.71 &0.56 &- \\
    \cline{2-8}
    
    &\multirow{2}{*}{XVO~\cite{lai2023xvo} }
    &$t_{\text{rel}}$ &3.92 &5.96 &- &{3.31} &- \\
    & &$r_{\text{rel}}$ &\tu{1.46} &3.96 &- &1.52 &- \\
    \cline{2-8}
    &\multirow{2}{*}{DualRefine~\cite{bangun2023dualrefine}} 
    &$t_{\text{rel}}$ &- &- &3.43 &6.80  &- \\
    & &$r_{\text{rel}}$ &- &- &1.04 &{1.13} &- \\
    \cline{2-8}

    % https://ieeexplore.ieee.org/stamp/stamp.jsp?tp=&arnumber=10568572
    &\multirow{2}{*}{ALH-VO$^\mp$~\cite{liu2024alhvo} }
    &$t_{\text{rel}}$ &\tu{3.64} &\tu{\tb{2.53}} &{2.01} &2.01 &\tu{2.55} \\
    & &$r_{\text{rel}}$ &1.88 &\tu{\tb{1.46}} &1.35 &1.32 &\tu{1.50} \\
    \cline{2-8}

    \hline
    \multirow{10}{*}{\rotatebox{90}{\parbox{70pt}{\centering General \hspace{2cm} Frame-to-Frame VO}}}
    &\multirow{2}{*}{TartanVO~\cite{wang2021tartanvo}}
    &$t_{\text{rel}}$ &4.72 &4.32 &6.00 &6.89 &5.48 \\
    & &$r_{\text{rel}}$ &2.95 &3.41 &3.11 &2.73 &3.05 \\
    \cline{2-8}
    
    &\multirow{2}{*}{DiffPoseNet~\cite{parameshwara2022diffposenet}}
    &$t_{\text{rel}}$ &2.93 &4.06 &4.02 &3.95 &3.74 \\
    & &$r_{\text{rel}}$ &1.76 &2.35 &\tu{0.51} &1.23 &1.46 \\
    \cline{2-8}
    
    &\multirow{2}{*}{iSLAM-VO$^\mp$~\cite{fu2024islam}}
    &$t_{\text{rel}}$ &4.30 &3.23 &4.22 &4.63 &4.09 \\
    & &$r_{\text{rel}}$ &1.16 &\underline{1.75} &1.41 &1.40 &\tu{\tb{1.43}} \\
    \cline{2-8}

    &\multirow{2}{*}{MAC-VO$^\mp$~\cite{qiu2025macvo}}
    &$t_{\text{rel}}$ &2.91 &3.56 &6.89 &8.03 &5.34 \\
    & &$r_{\text{rel}}$ &1.74 &2.58 &2.69 &3.82 &2.70 \\
    \cline{2-8}
    
    &\multirow{2}{*}{\textbf{Ours (DINO-VO)} }
    &$t_{\text{rel}}$ &\tu{\tb{1.65}}   &\underline{2.57}   &\tu{\tb{1.64}} &\tu{\tb{0.96}} &\tu{\tb{1.70}} \\
    & &$r_{\text{rel}}$ &\tu{\tb{0.82}} &3.31               &0.93           &\tu{0.71} &{1.44} \\
    \hline
    \end{tabularx}
    {\parbox{1.0\columnwidth}{- $t_{rel}$ ($\%$) and $r_{rel}$ (deg/100m) are the average translational and rotational RMSE drift on various segments with length 100-800 m.}} \\
    {\parbox{1.0\columnwidth}{- The best of each driving scenarios and general frame-to-frame VO results are \tu{underlined}. The best overall results are \tb{bold}.}} \\
    {\parbox{1.0\columnwidth}{- $^\mp$ The method uses stereo camera setup.}}
    \vspace{-0.3cm}
\end{table}

\begin{table}[t!]
    \centering
    
    \caption{\vspace{-2pt}Generalization Assesment Against other General SLAM/VO methods on KITTI Odometry Sequence 0-10 using ATE[m]$\downarrow$.}
    \label{tab:kitti_ate}
    \vspace{-6pt}
    \setlength{\tabcolsep}{3.0pt}
    \begin{tabularx}{\columnwidth}{c|cccc|cccccc}
    \hline
    &\multicolumn{4}{c|}{\centering \tb{SLAM}} &\multicolumn{6}{c}{\centering \tb{VO}} \\ 
    \rott{Sequence} &\rott{ORB-SLAM3\cite{campos2021orb3}} &\rott{LDSO~\cite{gao2018ldso}} &\rott{DPV-SLAM++\cite{lipson2024dpvslam}} &\rott{MambaVO++\cite{wang2025mambavo}} &\rott{SVO~\cite{forster2016svo}} &\rott{DSO~\cite{engel2017dso}} &\rott{DROID-VO~\cite{teed2021droid}} &\rott{DPVO\cite{teed2024dpvo}} &\rott{MambaVO~\cite{wang2025mambavo}} &\rott{\tb{Ours \hspace{0.4cm} (DINO-VO)}} \\

    \hline
    %   ORB     LDSO DPVSLAM MAMBAVO++ SVO     DSO   DROID   DPVO    MAMBAVO    DINO
    00 &6.77  &9.32  &8.30   &\tu{\tb{6.19}}   &X  &126.7  &98.4  &113.2   &112.4   &\tu{39.8}\\
    01 &X     &11.68  &11.3   &\tu{\tb{8.04}}   &X  &165.0  &84.2  &12.7    &\tu{8.16}     &24.8\\
    02 &30.5  &31.98  &39.6   &\tu{\tb{27.73}}  &X  &138.7  &108.8 &123.4   &93.78    &\tu{42.6}\\
    03 &\tu{\tb{1.04}}  &2.85  &2.50   &1.94   &X  &4.77   &2.58  &2.09    &1.80     &\tu{1.51}\\
    04 &0.93  &1.22  &0.78   &\tu{0.59}   &58.4  &1.08   &0.93  &0.68    &0.66     &\tu{\tb{0.35}}   \\
    05 &5.54  &5.1  &5.74   &\tu{\tb{3.05}}   &X  &49.5   &59.2  &58.9    &56.51    &\tu{15.4}   \\
    06 &16.6  &13.55  &\tu{11.6}   &11.79  &X  &113.6  &64.4  &54.8    &57.19    &\tu{\tb{3.91}}   \\
    07 &9.70  &2.96  &\tu{\tb{1.52}}   &1.7    &X  &27.9  &24.2  &19.2    &17.9     &\tu{7.39}   \\
    08 &\tu{60.7}  &129.0  &110.9  &105.4 &X  &120.2 &64.5  &115.9   &116.0   &\tu{\tb{21.6}}   \\
    09 &\tu{7.89}  &21.64  &76.7   &63.24  &X  &74.3   &71.8  &75.1    &73.56    &\tu{\tb{7.16}}   \\
    10 &\tu{8.65}  &17.36  &13.7   &10.51  &X  &16.3   &16.9  &13.6    &14.37    &\tu{\tb{1.31}}   \\
    \hline
    Avg &X    &22.42  &25.7   &\tu{21.84}  &X  &76.3   &54.2  &53.6    &50.21    &\tu{\tb{15.1}}\\

    \hline
    \end{tabularx}
    {\parbox{1.0\columnwidth}{- The best of each SLAM and VO results are \tu{underlined}. The best overall results are \tb{bold}.}}
    \vspace{-0.3cm}
\end{table}

\subsection{Generalization to Outdoor Driving Scenarios}
\label{sec:generalization}
We evaluate the generalization ability of DINO-VO on KITTI dataset~\cite{geiger2013visionkitti} which contains outdoor driving sequences, featuring dynamic objects such as cars and pedestrians which provides significant domain gap with TartanAir dataset.
In addition to comparing with general VO methods, we also provide the comparison with frame-to-frame VO methods which are specifically trained and/or designed for outdoor driving scenarios~\cite{wang2017deepvo, zhan2020dfvo, liu2022hdvo, lai2023xvo, bangun2023dualrefine, liu2024alhvo}.

In Table \ref{tab:kitti_rpe}, we provide comparison with other frame-to-frame VO methods using RMSE drift~\cite{geiger2013visionkitti}. 
Compared to the general methods, DINO-VO achieves the lowest $t_{rel}$ on 3 out of 4 sequences and the lowest $r_{rel}$ on 2 out of 4 sequences.
On average, DINO-VO achieves the lowest $t_{rel}$ with 54\% error reduction compared to the second best results, DiffPoseNet.
Our method achieves the second lowest average $r_{rel}$ with less than 1\% increase compared to iSLAM-VO.
Despite never having seen these scenarios, DINO-VO provides significant improvement against general frame-to-frame VO methods~\cite{wang2021tartanvo,parameshwara2022diffposenet,fu2024islam, qiu2025macvo} and even methods which are designed/trained for driving scenarios \cite{wang2017deepvo, zhan2020dfvo, liu2022hdvo,lai2023xvo, bangun2023dualrefine, liu2024alhvo}, showcasing its generalization ability.

In Table \ref{tab:kitti_ate}, we compare DINO-VO with other general multi-frame VO/SLAM methods in term of ATE on KITTI odometry sequence 0-10, specifically reported in \cite{lipson2024dpvslam}. 
DINO-VO has the lowest ATE in 5 out of 11 sequences and achieves the lowest average ATE despite not using multi-frames optimization and loop-closure.
Note that on sequence 00, 02, 05, and 07, the visual SLAM systems benefit from loop-closure resulting in better accuracy.
On Sequence 01, 03, 06, 08, 09, and 10 which contain significant appearances of dynamic objects, DINO-VO achieves the lowest ATE on 4 out of 6 sequences despite having never seen the objects and using no explicit matches filtering, showcasing its robustness to dynamic objects.

\begin{figure}[t!]
\centering
\includegraphics[width=1.0\columnwidth]{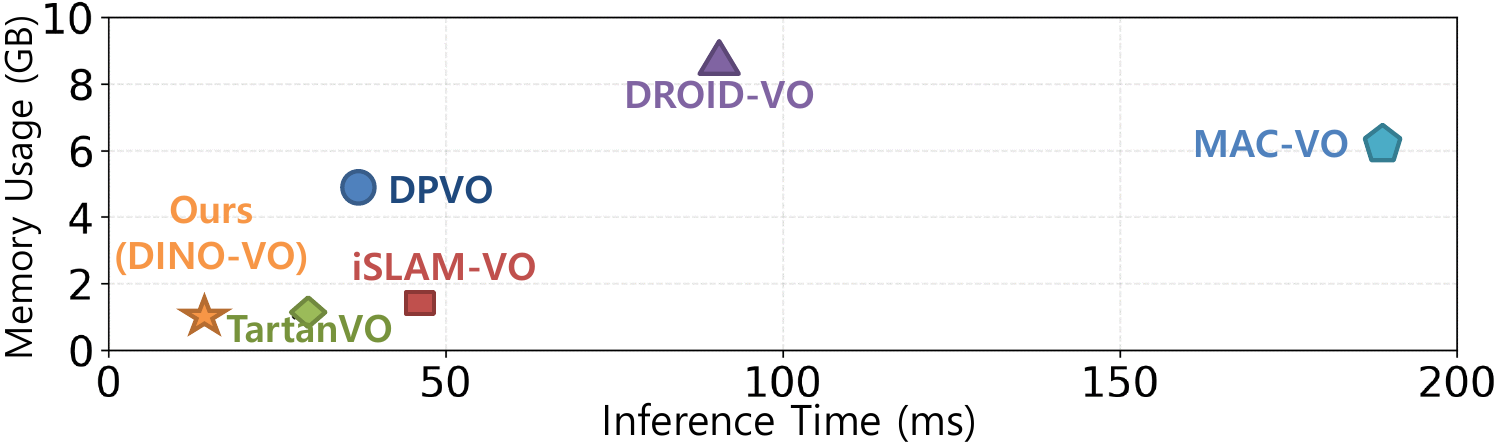}

\vspace{-0.3cm}
\caption{Inference time and memory usage comparison on EuRoC sequences using a single RTX4090 GPU and i9-12900k CPU. DINO-VO uses 476$\times$742 image input and half-precision inference (FP16) while other methods follow their default implementation.}
\vspace{-0.3cm}
\label{fig/runtime_comparison}
\end{figure}

\begin{table}[t!]
    \centering
    \caption{Timing Breakdown of DINO-VO.}
    \label{tab:runtime}
    \vspace{-6pt}
    \begin{tabularx}{\columnwidth}{cccccc}
    \hline
    $\operatorname{t_{Det}}$ &$\operatorname{t_{FinerCNN}}$ &$\operatorname{t_{DINOv2}}$ &$\operatorname{t_{Match}}$ &$\operatorname{t_{Pose}}$ &$\operatorname{t_{Total}}$\\
    \hline
    0.65 ms &1.10 ms &4.05 ms &3.85 ms &4.15 ms &13.8 ms\\
    \hline
    
    \end{tabularx}
    \vspace{-0.4cm}
\end{table}

\subsection{Runtime and Memory}
\label{sec:runtime_and_memory}
Fig. \ref{fig/runtime_comparison} shows the inference time and memory comparison of DINO-VO with other methods on EuRoC dataset using a single RTX4090 GPU and i9-12900k CPU.
DINO-VO runs at 13.8 ms (~72 FPS) and uses only 0.96GB GPU memory.
All of the reported DINO-VO results are run using FP16 half-precision inference with PyTorch software.
DINO-VO has up to 9$\times$ and 13.5$\times$ lower memory footprint and faster inference time, respectively.
Compared to prior works, DINO-VO uses sparse feature matching instead of optical-flow~\cite{teed2021droid, teed2024dpvo, wang2021tartanvo, fu2024islam, qiu2025macvo}.
Furthermore, DINO-VO benefits from highly efficient parallelization of attention mechanism in its feature descriptor~\cite{dosovitskiy2021vit} and feature matching networks~\cite{lindenberger2023lightglue}.
We further provide the timing breakdown of DINO-VO in Table \ref{tab:runtime}.
The proposed keypoints detector and fine-grained CNN encoder have minimal overhead compared to the total inference time, using only 5\% and 8\% of the total runtime.

\begin{table}[t!]
    \centering
    \caption{Ablation Study Result.}
    \label{tab:ablation}
    \setlength{\tabcolsep}{4.0pt}
    \vspace{-6pt}
    \begin{tabularx}{\columnwidth}{cc|cc|ccc}
    \hline
    &{ } &\multicolumn{2}{c|}{\parbox{3cm}{\centering \tb{Matching Performance \\ On TartanAir}}} &\multicolumn{3}{c}{\tb{Traj Eval.}} \\
    &Det / Desc 
    &{\parbox{0.4cm}{\centering R}}
    &{\parbox{2.4cm}{\centering MMA @1/3/5/10 px}}
    &\rots{TartanAir ATE[m]$\downarrow$} 
    &\rots{EuRoC ATE[m]$\downarrow$} 
    &\rots{KITTI ATE[m]$\downarrow$}\\
    \hline
    \multirow{4}{*}{(1)} 
    &SP / SP &\tb{92.7} &\tu{36.8} / 70.4 / 78.3 / \tu{83.3} &1.14 &\tu{0.616} &\tu{28.2} \\
    &SP / D  &89.1       &36.4 / 69.1 / 76.4 / 80.7 &\tu{1.03} &0.630 &56.1 \\
    &SP / F  &\tu{92.6} &36.7 / \tu{71.0} / \tu{78.6} / \tu{83.3} &1.11 &0.634 &30.3\\
    &SP / D+F &92.2 &\tb{38.0} / \tb{73.0} / \tb{80.9} / \tb{84.6} &\tb{0.68} &\tb{0.420} &\tb{20.1}\\
    \hline
    \multirow{5}{*}{(2)} 
    &SP / D+F &\tu{92.2} &\tb{38.0} / \tb{73.0} / {80.9} / \tu{84.6} &\tu{0.68} &\tu{0.420} &20.1\\
    &SP-A / D+F &\tb{92.6} &\tu{36.9} / 71.3 / 80.1 / 83.3 &0.71 &0.460 &28.1 \\
    &SIFT / D+F &83.4 &33.8 / 67.0 / 75.1 / 77.3 &1.14 &0.584 &25.9 \\
    &SIFT-A / D+F  &88.7 &31.8 / 65.6 / 78.6 / 83.7 &0.92 &0.565 &\tu{18.2}\\
    &SAL / D+F &88.3 &33.9 / \tu{71.6} / \tu{82.2} / 83.6 &0.78 &0.528 &{19.8} \\
    &SAL-A / D+F &91.0 &33.1 / 68.5 / \tb{82.8} / \tb{87.0} &\tb{0.58} &\tb{0.404} &\tb{15.1} \\
    \hline
    \end{tabularx}
    {\parbox{1.0\columnwidth}{- The best and second best results of each setting (1) and (2) are \tb{bold} and \tu{underlined}. }} \\
    {\parbox{1.0\columnwidth}{- SP: SuperPoint (baseline); D: DINOv2; F: FinerCNN, D+F: Concatenated DINOv2 and FinerCNN (Proposed in Sec. \ref{sec:feat_desc}).}} \\
    {\parbox{1.0\columnwidth}{- SP-A: SuperPoint Detector w/ grid alignment; SIFT-A: SIFT detector w/ grid-alignment; SAL: Salient Detector w/o grid alignment; SAL-A: SAL w/ grid aligment (Proposed in Sec. \ref{sec:keypoint_det}).}}
    \vspace{-0.4cm}
\end{table}

\section{Ablation Studies}
\label{sec:ablation}
\begin{figure*}[t!]
\begin{center}
\includegraphics[width=1.0\textwidth]{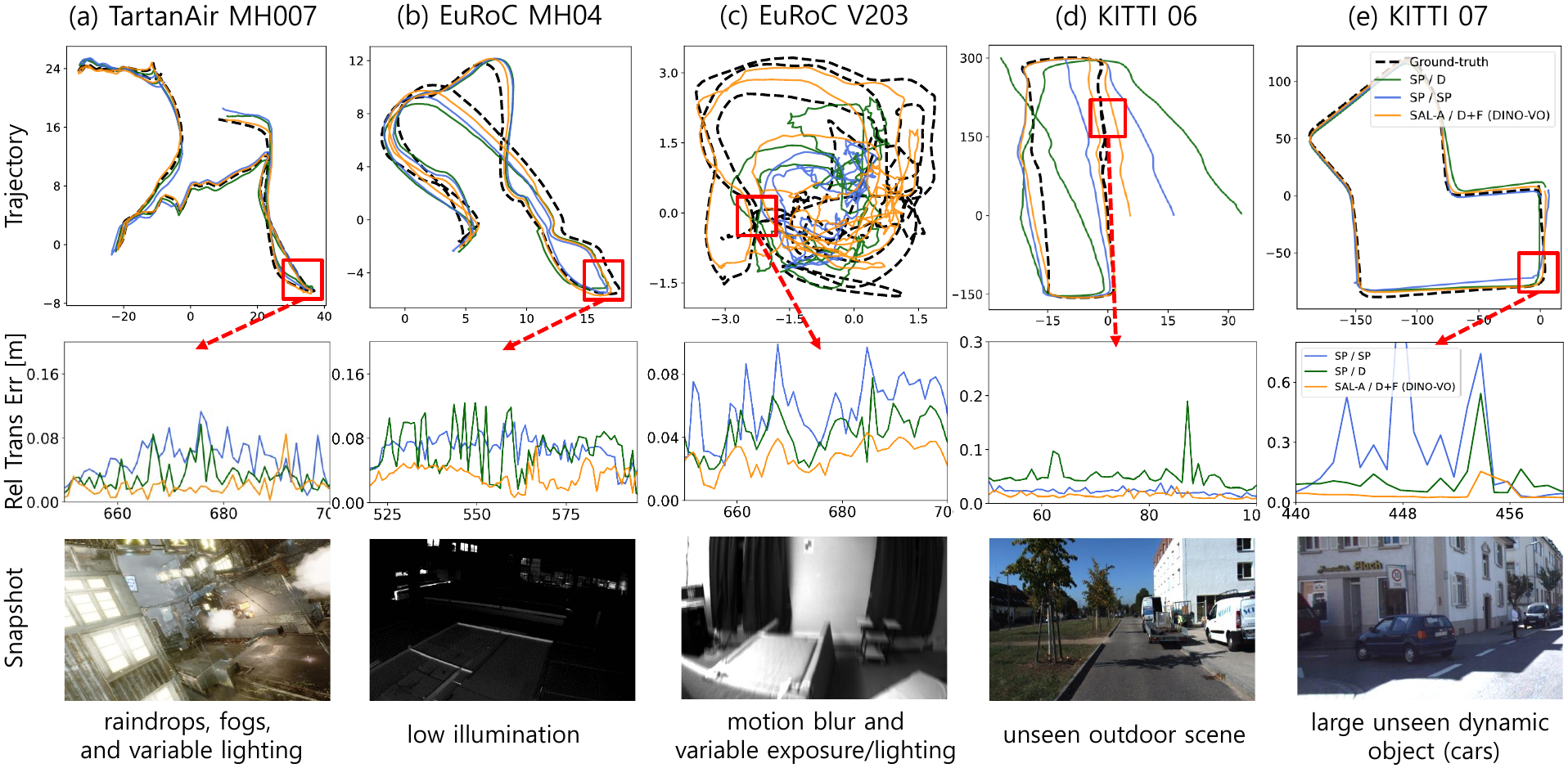}
\end{center}
\vspace{-0.3cm}
\caption{Qualitative results. \textbf{top row}: Trajectory comparisons of the proposed method \textbf{(DINO-VO}) along with two ablated systems, namely ours with \textbf{SP / SP} and ours with \textbf{SP / D}. \textbf{middle row}: frame-to-frame relative translation error comparison on the red square area of the trajectory. \textbf{bottom row}: snapshot of the image input along with its challenging conditions which disturb the VO systems. DINO-VO has the lowest error on various challenging conditions compared to with the system with SuperPoint detector-descriptor and DINOv2 only features, showcasing the robustness of our system.}
\vspace{-0.3cm}
\label{fig/traj_result}
\end{figure*}

To evaluate the effectiveness of our keypoint detectors and feature descriptors, we conduct ablation studies with different settings. 
In setting (1), we built on top of SuperPoint (SP) joint detector-descriptor networks\cite{detone2018superpoint} as the baseline and ablate the feature descriptors.
In setting (2), we ablate the keypoints detectors on top of setting (1) SP/D+F.
*-A denotes keypoints detector with grid-alignment proposed in Sec. \ref{sec:keypoint_det}.
We report the matching performance in term of recall at 5 pixel error threshold and mean matching accuracy (MMA) at various error thresholds\cite{zhan2024imatching} as well as the average ATE.
The ablation study results are shown in Table \ref{tab:ablation}.
We further depict the qualitative comparisons in Fig. \ref{fig/traj_result}.

\subsection{Localizable Features Importance}
As shown in Table \ref{tab:ablation}, using DINOv2 descriptors alone (D) results in matching performance drop compared to using SuperPoint descriptors, similar results are observed in \cite{jiang2024omniglue}.
The combination of DINOv2 with FinerCNN (D+F) significantly improves the matching accuracies across all thresholds compared to SuperPoint descriptors. 
These results show the importance of having localizable features from FinerCNN.
The FinerCNN alone has comparable matching performance to SuperPoint while having a 2.4$\times$ faster inference speed.
Beyond the matching performance, the proposed descriptors achieve 40\%, 32\% and 29\% average ATE reduction on TartanAir, EuRoC, and KITTI, compared to using SuperPoint features.

\subsection{Robustness and Generalization of DINOv2 Features}
\label{sec:ablation_robustness}
As provided in Table \ref{tab:ablation}, the usage of DINOv2 without FinerCNN shows better accuracy on TartanAir MH sequences which contain significant disturbances\cite{wang2020tartanairdataset} despite having less accurate matches.
The semantic information provided by DINOv2 is observed to be beneficial to down-weight unfavorable matches for pose estimation.
However, using DINOv2 features alone performs significantly worse when tested on out-of-domain sequences as shown in Table \ref{tab:ablation} and Fig. \ref{fig/traj_result}. (d)-(e).
We suspect that the matching networks learns to exploit specific bias of DINOv2 features in the TartanAir dataset hindering the generalization ability of DINOv2 features.
Combining with FinnerCNN resolves this issue as the matching network can optimally combine the geometric and semantic information encoded in FinerCNN and DINOv2 features.
This combination further improve the robustness which are pronounce in sequences with various disturbances such as variable lighting, fogs, aggressive motion, motion blur and unseen dynamic object as shown Fig. \ref{fig/traj_result}. (a), (c), and (e).

\subsection{Keypoints Detector Effectiveness}
The proposed keypoints detector (SAL-A) used in DINO-VO significantly improves the matching and VO accuracy compared to using SuperPoint and SIFT detectors.
Without grid-alignment, salient detector (SAL) performs worse compared to SuperPoint detector, showing the importance of having aligned keypoints on DINOv2 coarse features.
Introducing grid-alignment to SIFT detector (SIFT-A) significantly improves its matching and VO performance.
Conversely, SuperPoint detector w/ alignment (SP-A) results in a slightly degraded performance.
Furthermore, SAL-A offers more distributed keypoints across the image which is beneficial for pose estimation.
Additionally, it has 6$\times$ and 4$\times$ faster detector inference time compared to SIFT and SP.

\section{Discussion And Conclusion}
\label{sec:conclusion}
In this paper we present DINO-VO, a novel feature-based visual odometry approach that leverages the robustness and generalization capabilities of the DINOv2 visual foundation model. 
We verify that DINO-VO is robust against various disturbances and generalizable when tested on out-of-domain datasets. 
DINO-VO achieves state-of-the-art accuracy against previous frame-to-frame VO methods on the TartanAir and KITTI datasets, while also demonstrating highly competitive performance on the EuRoC dataset.
DINO-VO is designed to be lightweight and able to run at 72 FPS with less than 1GB memory usage on a single GPU.
In addition, we have addressed the problem of directly utilizing DINOv2 coarse features in sparse image matching as highlighted in \cite{jiang2024omniglue}.

The main limitation of our approach is that it relies on frame-to-frame pose estimation such that it is more prone to drift compared to multi-frame VO methods. 
Nevertheless, our monocular method is closing the accuracy gap with multi-frame VO methods on TartanAir and EuRoC, and remains highly competitive even against top-performing stereo systems like MAC-VO when considering its significant advantages in efficiency and memory usage.
Noticeably, our method outperforms multi-frames VO methods on KITTI dataset. 
Future work will investigate integrating DINO-VO on multi-frames VO schemes such as using bundle adjustment. 
These improvements could also benefits from metric depth estimation developed in \cite{yang2024depth} which also utilizes DINOv2 encoder. 


%%%%%%%%%%%%%%%%%%%%%%%%%%%%%%%%%%%%%%%%%%%%%%%%%%%%%%%%%%%%%%%%%%%%%%%%%%%%%%%%

%%%%%%%%%%%%%%%%%%%%%%%%%%%%%%%%%%%%%%%%%%%%%%%%%%%%%%%%%%%%%%%%%%%%%%%%%%%%%%%%



% \section*{ACKNOWLEDGMENT}

% \clearpage 
\bibliographystyle{ieeetr}
\bibliography{root}

\end{document}